\documentclass[sigconf]{acmart}

\AtBeginDocument{%
  \providecommand\BibTeX{{%
    Bib\TeX}}}

\usepackage{algorithm}
\usepackage{algorithmic}

\usepackage{times}
\usepackage{soul}
\usepackage{url}
\usepackage[utf8]{inputenc}
\usepackage{graphicx}
\usepackage{amsmath}
\usepackage{amsthm}
\usepackage{booktabs}
\usepackage[switch]{lineno}
\usepackage{subfigure}
\usepackage{subcaption}
\usepackage{multirow}
\usepackage{diagbox}
\usepackage{makecell}
\usepackage{comment}
\usepackage{afterpage}
\usepackage{soul}

\setcopyright{acmlicensed}
\copyrightyear{2018}
\acmYear{2018}
\acmDOI{XXXXXXX.XXXXXXX}

\acmConference[CODS-COMAD Dec '24]{8th International Conference on Data Science and Management of Data (12th ACM IKDD CODS and 30th COMAD)}{December 18--21, 2024}{Jodhpur, India}
\acmBooktitle{8th International Conference on Data Science and Management of Data (12th ACM IKDD CODS and 30th COMAD) (CODS-COMAD Dec '24), December 18--21, 2024, Jodhpur, India}
\acmISBN{978-1-4503-XXXX-X/18/06}

\acmSubmissionID{149}

\begin{document}

\title {Framework for Co-distillation Driven Federated Learning to Address Class Imbalance in Healthcare}
\author{Suraj Racha}
\authornote{Both authors contributed equally to this research.}
\affiliation{%
  \institution{Indian Institute of Technology, Bombay}
 \city{Mumbai}
 \country{India}
 }
\email{23d1627@iitb.ac.in}

\author{Shubh Gupta}
\authornotemark[1]
\affiliation{%
  \institution{Sardar Patel Institute of Technology}
 \city{Mumbai}
 \country{India}
 }
\email{shubh.gupta@spit.ac.in}

\author{Humaira Firdowse}
\affiliation{%
  \institution{Indian Institute of Technology, Madras}
 \city{Chennai}
 \country{India}
 }
\email{humairafirdowse@gmail.com}

\author{Aastik Solanki}
\affiliation{%
  \institution{Indian Institute of Technology, Bombay}
 \city{Mumbai}
 \country{India}
 }
\email{aastik.solanki99@gmail.com}

\author{Ganesh Ramakrishnan}
\affiliation{%
  \institution{Indian Institute of Technology, Bombay}
 \city{Mumbai}
 \country{India}
 }
\email{ganesh@cse.iitb.ac.in}

\author{ Kshitij S. Jadhav}
\affiliation{%
  \institution{Indian Institute of Technology, Bombay}
 \city{Mumbai}
 \country{India}
 }
\email{kshitij.jadhav@iitb.ac.in}

\renewcommand{\shortauthors}{Racha, Gupta et al.}

\title [Co-distillation for Federated Learning in Healthcare]{Framework for Co-distillation Driven Federated Learning to Address Class Imbalance in Healthcare}
\begin{abstract}
  Federated Learning (FL) is a pioneering approach in distributed machine learning, enabling collaborative model training across multiple clients while retaining data privacy. However, the inherent heterogeneity due to imbalanced resource representations across multiple clients poses significant challenges, often introducing bias towards the majority class. This issue is particularly prevalent in healthcare settings, where hospitals acting as clients share medical images. To address class imbalance and reduce bias, we propose a co-distillation driven framework in a federated healthcare setting. Unlike traditional federated setups with a designated server client, our framework promotes knowledge sharing among clients to collectively improve learning outcomes. Our experiments demonstrate that in a federated healthcare setting, co-distillation outperforms other federated methods in handling class imbalance. Additionally, we demonstrate that our framework has the least standard deviation with increasing imbalance while outperforming other baselines, signifying the robustness of our framework for FL in healthcare.
\end{abstract}




\maketitle

\section{Introduction}
Federated learning (FL) \cite{FL} has been a significant approach for collaborative model training while catering to privacy preserving. In a resource-constrained and privacy-sensitive domain like healthcare, FL becomes of great relevance where data can still remain at client level while model training continues. Hospitals, viewed as clients, are less likely to possess sufficient and balanced data for training. However, they can benefit from a collaborative approach to train models.
One of the foremost challenges is the heterogeneity inherent in the data of each client due to different data collection techniques and client-specific preferences. Each hospital may have a different number of patients diagnosed with separate diseases based on their specialized areas of care. In such an imbalanced scenario, models may be biased towards the majority class, while under-performing in representing the minority class.
In our work, we propose to mitigate effects of severe class imbalance on the prediction of minority class in medical domain. This is done using a co-distillation approach for FL. In this method, clients actively exchange knowledge during the training process, by sharing probabilistic predictions or soft labels. These shared soft labels serve to enhance the model's confidence, ultimately increasing its accuracy while still preserving privacy as no feature representations or data is shared. Moreover, since co-distillation is a knowledge distillation approach based on mutual sharing of knowledge through soft targets and not model parameters, it experiences reduced communication overhead compared to other FL baselines that communicate local model updates in each iteration\cite{wu2022communication, li2024federated}. We release our codes at \url{https://github.com/humairafirdowse/codistillation}.

We summarize the contributions of our work as follows: 
(1) We demonstrate the superior performance of a co-distillation based federated learning framework in addressing severe class imbalance in healthcare settings. 
(2) We highlight the robust performance of co-distillation compared to other baselines by reporting the lowest standard deviation in results across varying skew levels. Additionally, our framework achieves superior outcomes even with varying number of clients.
(3) We further demonstrate that our framework excels in resource-constrained settings where the total number of labelled medical images for model training is limited.

\begin{figure*}[t]
\centering
  \includegraphics[width=2\columnwidth]{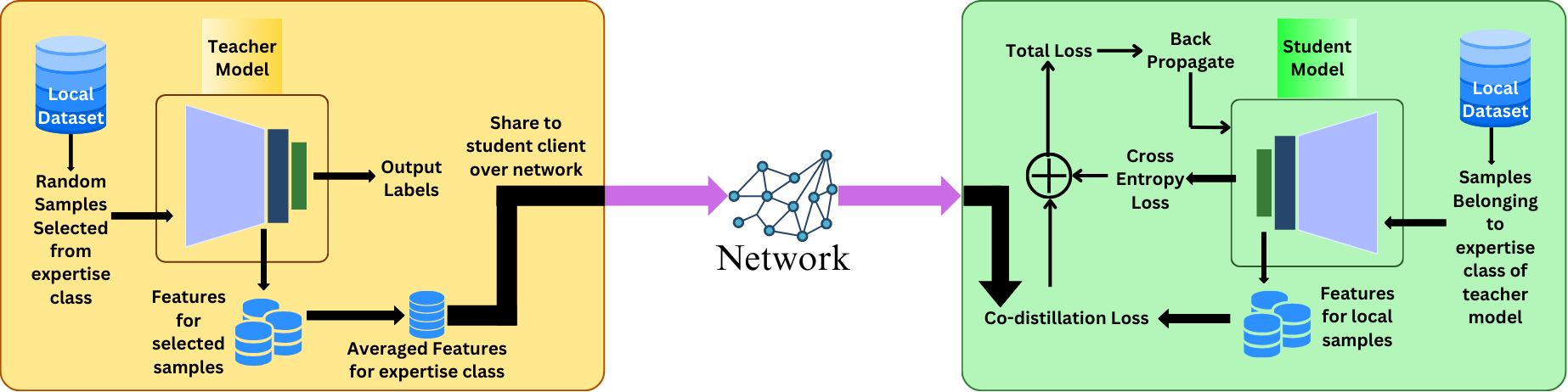}
  \caption{Student Client randomly select another client as a teacher which shares averaged representations of their expertise class, while the student client uses these representations for knowledge distillation}
  \label{fig1}
\end{figure*}
\vspace{-0.5em}

\section{Related Work}
\textbf{Class Imbalance in Healthcare}: Class imbalance poses significant challenges in healthcare where distribution of classes are skewed. Prior works like GAN and SMOTE have been used to address the same\cite{goodfellow2014generative, chawla2002smote}. However, artificial images pose ethical concerns with sensitive data like medical information\cite{paladugu2023generative}. Previous studies have also leveraged meta-learning and model tuning to address imbalance. Recent works also developed FL architectures using SVM and distributed learning for medical domain\cite{brisimi2018federated, lu2019learn}.
\\
\textbf{Class Imbalance in Federated Settings}:
AutoBalance \cite{li2021autobalance} dynamically designs a training loss function to enhance accuracy and fairness by personalizing treatment. Others \cite{wang2021addressing} propose a novel monitoring scheme and "Ratio Loss" function to mitigate class imbalance. GAN-based\cite{gan} and SMOTE-based\cite{younis2022fly} FL frameworks like BalanceFL\cite{balancefl} and FEDGAN-IDS\cite{fedganids} allow synthetic data generation at the client level to better represent minority classes.
Knowledge distillation (KD) \cite{10258226} in FL facilitates the transfer of knowledge from a complex model (teacher) to a simpler model (student), enhancing performance and efficiency. Federated Distillation (FD) addresses heterogeneity by supporting diverse model architectures, adapting to varying system capabilities, and facilitating knowledge transfer to overcome data distribution discrepancies, thereby enhancing performance \cite{li2024federated}. FedKD \cite{fedkd} leverages KD to reduce communication, enabling efficient and privacy-preserving deployments in  healthcare. FedTweet \cite{fedtweet} employs a novel "Two-fold Knowledge Distillation" approach to address non-IID data challenges. Previous works like BalanceFL and FEDGAN-IDS also use GAN-based augmentation to handle long-tailed data. 

\begin{algorithm}
\caption{Co-distillation - Teacher client side\protect\footnotemark.}
\begin{algorithmic}[1]
\label{algorithm1}
\REQUIRE Expertise class $c$ of teacher client, number of samples $k$
\STATE $\chi \leftarrow \text{Sample } k \text{ data points from } X \text{ with } y = c$
\STATE $R \leftarrow \text{AVG}(\text{feature}(\chi))$
\RETURN $R$
\end{algorithmic}
\end{algorithm}
\footnotetext{AVG: Average of all features}

\begin{algorithm}
\caption{Co-distillation - Student client Side \protect\footnotemark}
\begin{algorithmic}[1]
\label{algorithm2}
\REQUIRE Number of rounds $N_r$, Inputs $X$, Targets $Y$, co-distillation coefficient $\lambda$
\FOR{$i \leftarrow 1$ \TO $N_r$}
    \STATE $T \leftarrow \text{Select random client as Teacher}$
    \STATE $c \leftarrow \text{Expertise class of the Teacher client T}$
    \STATE $R \leftarrow \text{Average representation for class } c \text{ by client } T$
    \STATE $loss = \text{CE}(Y, M(X))$
    \FORALL{$x, y \in X, Y$ \textbf{where} $y = c$}
        \STATE $loss = loss + \lambda \times \text{MSE}(\text{feature}(x), R)$
    \ENDFOR
    \STATE Back-propagate
\ENDFOR
\end{algorithmic}
\end{algorithm}
\footnotetext[2]{CE: Cross Entropy Loss and MSE: Mean Squared Error}

\section{Methodology}
\subsection{Aim}
We aim to show the effectiveness of co-distillation for minority class prediction in high skew or imbalance conditions. A total of $N_c$ personalized models are trained for $N_c$ total clients respectively. Final model at each client are then used to predict the labels of minority class specific to that client.
\begin{table*}[ht]
  \begin{minipage}{0.48\textwidth}
    \centering
    \resizebox{\linewidth}{!}{%
      \begin{tabular}{@{}llllll|lllll@{}}
        \toprule
        \multirow{3}{*}{\textbf{Method}} & \multicolumn{5}{c|}{\textbf{skew \% (APTOS)}} & \multicolumn{5}{c}{\textbf{skew \% (COVID-19)}} \\ \cmidrule(l){2-11}
        & 0.00 & 20.0 & 40.0 & 60.0 & sd & 0.0 & 20.0 & 40.0 & 60.0 & sd \\
        \cmidrule(r){1-1} \cmidrule(l){2-11} CD 
        & 88.1 & \textbf{95.9} & 83.4 & \textbf{81.7} & \textbf{0.06} & \textbf{95.4} & \textbf{95.8} & \textbf{89.1} & \textbf{90.3} & \textbf{0.03} \\
        FD & \textbf{93.5} & 89.8 & 83.1 & 69.4 & 0.11 & 89.5 & 85.0 & 73.2 & 48.3 & 0.18 \\
        FP & 92.7 & 90.5 & \textbf{86.2} & 75.2 & 0.08 & 89.6 & 82.9 & 72.2 & 51.6 & 0.17 \\
        FAMP & 92.9 & 90.1 & 83.1 & 69.3 & 0.11 & 88.7 & 85.6 & 71.7 & 60.3 & 0.13 \\
        FAvg & 90.1 & 90.0 & 63.4 & 0.60& 0.42 & 92.9 & 0.00 & 0.00 & 0.00 & 0.46 \\ \bottomrule
      \end{tabular}%
    }
    \caption{Classification Accuracy (\%) on non-expertise class. Clients = $4$ and $600$ images for APTOS and $200$ images for COVID per each class for varying skew (\%). \textit{CD: Co-distillation, FD: FedDistill, FP: FedProto, FAMP: FedAMP, FAvg: FedAvg}}
    \label{tab:table1}
  \end{minipage}\hfill
  \begin{minipage}{0.48\textwidth}
    \centering
    \resizebox{\linewidth}{!}{%
      \begin{tabular}{@{}llllll|lllll@{}}
        \toprule
        \multirow{3}{*}{\textbf{Method}} & \multicolumn{5}{c|}{\textbf{skew \% (APTOS)}} & \multicolumn{5}{c}{\textbf{skew \% (COVID-19)}} \\ \cmidrule(l){2-11}
        & 0.00 & 20.0 & 40.0 & 60.0 & sd & 0.0 & 20.0 & 40.0 & 60.0 & sd \\ \cmidrule(r){1-1} \cmidrule(l){2-11} CD
        & 88.5 & 88.2 & \textbf{85.3} & \textbf{81.0} & \textbf{0.03} & \textbf{92.9} & \textbf{92.3} & \textbf{87.2} & \textbf{70.9} & \textbf{0.10} \\
        FD & 91.9 & 88.4 & 80.2 & 64.4 & 0.12 & 88.0 & 80.9 & 68.0 & 35.2 & 0.23 \\
        FP & \textbf{93.3} & \textbf{89.4} & 84.7 & 67.9 & 0.11 & 89.0 & 79.9 & 66.6 & 35.1 & 0.24 \\
        FAMP & 92.2 & 89.2 & 80.8 & 62.7 & 0.13 & 88.4 & 80.3 & 66.8 & 33.0 & 0.24 \\
        FAvg & 87.4 & 50.1 & 0.00 & 0.00 & 0.43 & 76.2 & 0.1 & 0.00 & 0.00 & 0.38 \\ \bottomrule
      \end{tabular}%
    }
    \caption{Classification Accuracy (\%) on non-expertise class. Clients = $6$ and $600$ images for APTOS and $200$ images for COVID per class) for varying skew (\%). \textit{CD: Co-distillation, FD: FedDistill, FP: FedProto, FAMP: FedAMP, FAvg: FedAvg}}
    \label{tab:table2}
  \end{minipage}
\end{table*}

\begin{table*}[ht]
\begin{minipage}{1\columnwidth}
\centering
\resizebox{1\columnwidth}{!}{%
\begin{tabular}{@{}llllll|llllll@{}}
\toprule
\multirow{2}{*}{\makecell{(skew = 0\%)\\\textbf{Method}}} & \multicolumn{5}{c|}{\textbf{Number of Images (APTOS)}} & \multicolumn{5}{c}{\textbf{Number of Images (COVID-19)}} \\ \cmidrule(l){2-11} 
& 200 & 300 & 400 & 500 & 600 & & 50 & 100 & 150 & 200 \\ \cmidrule(r){1-1}  \cmidrule(l){2-11} CD
& \textbf{92.8} & 87.6 & 88.4 & 83.1 & 88.1 & &84.6 &\textbf{ 94.5} & \textbf{93.7} & \textbf{95.4} \\
FD & 91.5 & 91.8 & \textbf{92.3 }& \textbf{92.3} & \textbf{93.5 }& &80.8 & 87.1 & 88.7 & 89.5 \\
FP & 92.3 & \textbf{ 92.8} & \textbf{92.3} & \textbf{92.3} & 92.7 & &81.0 & 88.4 & 89.0 & 89.6 \\
FAMP & 92.1 & 92.2 & 92.2 & 91.9 & 92.9 & &80.5 & 87.8 & 88.7 & 88.7 \\
FAvg & 86.2 & 88.6 & 87.6 & 87.4 & 90.1 & &\textbf{85.0} & 90.2 & 93.2 & 92.9 \\ 
\bottomrule
\end{tabular}%
}
\caption{Classification Accuracy (\%) on the non-expertise class with varying number of images in the APTOS and COVID-19 datasets for $\text{skew} = 0.00\%$. \textit{CD: Co-distillation, FD: FedDistill, FP: FedProto, FAMP: FedAMP, FAvg: FedAvg}}
\label{tab:my-table3}
\end{minipage}\hspace{1\columnsep}%
\begin{minipage}{1\columnwidth}
\centering
\resizebox{\columnwidth}{!}{%
\begin{tabular}{@{}llllll|lllllll@{}}
\toprule
\multirow{2}{*}{\makecell{(skew = 60\%)\\\textbf{Method}}} & \multicolumn{5}{c|}{\textbf{Number of Images (APTOS)}} & \multicolumn{5}{c}{\textbf{Number of Images (COVID-19)}} \\ \cmidrule(l){2-11} 
& 200 & 300 & 400 & 500 & 600 & &50 & 100 & 150 & 200 \\ \cmidrule(r){1-1} \cmidrule(l){2-11} CD
 & \textbf{85.9} & \textbf{ 78.9} & \textbf{80.5} & \textbf{84.1} & \textbf{81.7} & &\textbf{19.3} & \textbf{86.2} & \textbf{87.2} & \textbf{90.3} \\
FD & 57.8 & 60.6 & 63.9 & 63.8 & 69.4 & &3.52 & 20.7 & 43.8 & 48.3 \\
FP & 62.0 & 64.5 & 71.3 & 70.5 & 75.2 & &4.11 & 18.4 & 42.0 & 51.6 \\
FAMP & 55.0 & 60.6 & 63.5 & 65.3 & 69.3 & &2.35 & 16.9 & 44.4 & 60.3 \\
FAvg & 0.00 & 2.44 & 0.00 & 0.67 & 0.56 & &0.00 & 0.00 & 0.00 & 0.00 \\ 
\bottomrule
\end{tabular}%
}
\caption{Classification Accuracy (\%) on the non-expertise class with varying number of images in the APTOS and COVID-19 datasets for $\text{skew} = 60.0\%$. \textit{CD: Co-distillation, FD: FedDistill, FP: FedProto, FAMP: FedAMP, FAvg: FedAvg}}
\label{tab:my-table4}
\end{minipage}
\end{table*}

\subsection{Training via Co-Distillation}


Consider a FL setup with $N_c$ clients, where each client possess a local dataset. We initialize each client's model with a same base pretrained model prior to the actual training process. We define a client's expertise or majority class $c$ as the class with most data. 
In each training round, all clients act as individual student nodes and are trained in parallel. Each student node (client) randomly selects a teacher node from a list of available clients using uniform random sampling. This ensures that every other client has an equal chance of being chosen as a teacher, promoting diversity in the knowledge from clients with different areas of expertise. This randomness allows the student client to benefit from the varying data distributions of other clients, enhancing model robustness. Once the teacher client is selected, the student client requests soft targets specific to the teacher's expertise class $c$. Upon receiving this request, the teacher client samples a set of 'k' images (using uniform random sampling) corresponding to its expertise class $c$ and calculates the probabilistic outputs for these samples which are then transmitted to the student client over the network. Upon receiving the soft targets, the student client performs knowledge distillation by comparing the received soft targets with its own predictions for the samples from the same class $c$. In each round, clients update their local models based on both their own data and the distilled knowledge from the teacher client without any global model/server.

\vspace{-0.1cm}
\subsection{Student-Teacher Sharing}
The co-distillation process is governed by two algorithms: for the teacher client and student client, as shown in algorithm 1 and algorithm 2 respectively. At the teacher client, we take the expertise class $c$ of the teacher client and the number of samples $k$ as input parameters. We sample $k$ data points ($\chi$, using uniform random sampling) from the local dataset $X$ with the label $y = c$, followed by computation of soft targets on $\chi$.
For the student client, we take the number of rounds $N_r$, inputs $X$, targets $Y$, and the co-distillation coefficient $\lambda$ as inputs. At each round, we select a client as Teacher $T$ using uniform random sampling from other available clients, determine its expertise class $c$, and obtain the soft targets $R$ for class $c$ from client $T$. We then compute the primary Cross-Entropy loss, which measures the difference between the predicted logits and the target labels. This loss is updated with additional distillation loss through mean squared error between the student's local features and the received features from the teacher (represented by logits), weighted by $\lambda$. We then update the local model through back propagation based on this final loss (Figure \ref{fig1}).
Traditional FL methods may require full model parameter updates, while co-distillation focuses on transferring soft labels(logits). This approach significantly reduces the size of the information exchanged, which reduces the bandwidth and time required for communication. Importantly, only soft targets (logits) are shared between clients, ensuring that no raw data or direct features are exchanged.

\section{Experimental Setup}

\subsection{Setting}
\textbf{Datasets:} We have used APTOS \cite{aptos2019} and COVID-19 Radiography \cite{covid19radiography2020} (referred to as COVID hereafter) medical images dataset for evaluation. Both the datasets are taken from real world medical imaging data.
COVID is a binary class dataset with diseased and non-diseased labels, while APTOS originally consists of a normal class and four grades of diseased images. We have combined all the four diseased grades as a single disease class.
\\
\textbf{Design:} We design our experimental setting as follows: each local level of client sees class imbalance containing a major and a minor class. We define major or expert class as the class with maximum number of representative images. Consider the diseased and non-diseased binary classes as A and B, respectively. For a controlled environment of global balance, we choose half of the clients with A as majority and the other half with B as majority. Initially, we divide the total images of both classes equally among each client.
At $ \text{skew} = 0 $, each client has an equal number of images $ n_{A} $ for class $ A $ and $ n_{B} $ for class $ B $.
Imbalance denoted by $skew$ is applied on the minority class by randomly eliminating the skewed ratio portion. 
If the distribution at zero skew for a client is $ (n_A, n_B) $, the distribution at a particular percentage of $skew = s\%$ becomes $ (n_A, (1-0.01s)*n_B)$ when $B$ is the selected minority class of that client.

 

We evaluate our framework by subjecting training data to a skew of $\{0\%, 20\%, 40\%,$ and $60\%\}$ during experiments. We also vary the total number of training images available to represent a low resource setting where medical data is scarce or limited. The setting addresses effect of co-distillation when lesser training images are present during large imbalance. We consider the number of images per class  as $\{200, 300, 400, 500, 600\}$ for APTOS dataset and $\{50, 100, 150, 200\}$ for COVID dataset while keeping constant imbalance of $skew = 60\%$. To show the robustness of experiments, we implement the above in both 4-clients and 6-clients setting.


\subsection{Evaluation}
We use LeNet-5 \cite{lenet} architecture for each method with 3 convolution layers and 2 fully connected layers. We train all models for 100 epochs. 
We report the classification accuracy on minority class, \textit{i.e.}, $ \text{avg. accuracy} = \frac{\text{total minority class images correctly classified}}{\text{total minority class images}} $.
\\
\textbf{Baselines: } (i) \textbf{Communication-Efficient Learning of Deep Networks from Decentralized Data (FedAvg)} \cite{FL} is a distributed setting where averaged model parameters from each client is utilized~\cite{mcmahan2017communication}. (ii) \textbf{Personalized Cross-Silo Federated Learning on non-IID Data (FedAMP)} \cite{FedAMP} addresses heterogeneity challenge by facilitating pairwise collaboration between clients with similar data using attentive message passing. (iii) \textbf{Federated Prototype Learning across Heterogeneous Clients (FedProto)} aggregates local prototypes by sending the aggregated global prototypes back to all clients to regularize the training of local models ~\cite{tan2022fedproto}. (iv) \textbf{Federated Knowledge Distillation (FedDistill)} bears resemblance to the co-distillation approach, but with a notable difference in handling class representations. In FedDistill, average feature representations from all clients are utilized for a single class, irrespective of their expertise or non-expertise in these classes~\cite{seo202216}.

\section{Results and Discussions}
We report our minority class classification accuracy results using co-distillation and other FL baselines in a binary setup of diseased and non-diseased images. We also show the average standard deviation of accuracy on different methods. We report the average of three independent iterations for each experimental setup.


\subsection{Effects with increasing imbalance}
\label{sec:Variable data heterogeneity}
Tables \ref{tab:table1} and \ref{tab:table2} show the average classification accuracy with increasing skew/imbalance in both APTOS and COVID datasets for $clients = 4$ and $clients = 6$ respectively. In general, the minority class classification accuracy decreases with increasing skew. For APTOS dataset with $clients = 4$, co-distillation outperforms other baselines when a certain skew is introduced except at 40\% skew when FedProto sightly outperforms. Standard deviation is representative of the variance and robustness of a model's performance. Co-distillation has the least standard deviation of $0.06$. This shows that the performance drop is least with increasing imbalance for co-distillation. For COVID dataset we see that co-distillation outperforms all other baselines with a minimum standard deviation of $0.03$.
A similar trend can be observed when we increase number of clients to $clients = 6$ in table \ref{tab:table2}. For APTOS dataset, FedProto outperforms for no skew and 20\% skew while with further increasing skew, co-distillation has highest accuracy. Even with 6 clients, co-distillation performs best in all skews for COVID. It also has a significantly lesser standard deviation of $0.03$ and $0.10$ for APTOS and COVID respectively. The robustness is strongly witnessed at higher skews (eg. $skew = 60\%$) where other methods show significant performance decline. Few baseline methods like FedAvg completely loosing classification ability (\~ 0\% accuracy). Overall we demonstrate that co-distillation is more robust to data imbalance and maintains performance while other methods show significant decline.

\subsection{Effects with decreasing images}
\label{sec:Low resource settings}
Next, we address the question of \textit{performance in a high imbalance data setting when total data resources are limited}. Increasing skew degrades performance, with the impact being further intensified when training resources are limited. 
We study the effects of decreasing training images (limiting resources) with no skew and high skew (60\%).
As generally expected, the performance decreases on a high skew or imbalance. Table \ref{tab:my-table3} at no skew shows that all baselines perform well including FedAvg for both datasets, \textit{i.e.} there is no one best method. For no skew, however we notice that the performance not necessarily decreases with decreasing images; rather accuracy can fluctuate within a range. This trend can be seen for all the methods. It can possibly be attributed to the fact that image sizes at $skew = 0$, like 200 (minimum used for APTOS) and around 100 (for COVID) are sufficient for saturated model training, above which contribution of more images may not be explicitly visible always. However at a high skew of 60\% in table \ref{tab:my-table4}, model performance declines the least for co-distillation. Co-distillation outperforms with a very high margin over other FL methods in all of the settings. Low resource settings is defined as $\text{{$N_i$}} = 200$ and $\text{{$N_i$}} = 50$ for APTOS and COVID datasets respectively. 
Accuracy drops from $92.8\%$ to $85.9\%$ for APTOS (200 images) and $84.5\%$ to $19.3\%$ for COVID-19 dataset (50 images). This trend persists even when the total number of images increases (Tables \ref{tab:my-table4}). Notably, even under high skew, performance of co-distillation remains comparable to that in high images settings; while other FL baselines exhibit a significant decrease in accuracy when total number of images decreases.


\section{Conclusion}
Co-Distillation framework in healthcare shows higher performance in class imbalance settings. In realistic scenarios, data may be highly skewed and can be  subjected to limited shareable data resources. 
Co-distillation, as compared to other FL baselines, proves to be effective in learning class representations for the minority class in highly skewed medical image data in most of the experimental settings. It outperforms other baselines when subjected to high skew as well as extreme conditions of low available training data by achieving higher accuracy in minority class prediction. It also demonstrates least performance variance, making it a robust approach in addressing class imbalance settings in healthcare. 
Our current work shows majorly on data with two classes primarily to address diseased and non-diseased as category. However, future work can include experiments with more refined classes.
\bibliographystyle{ACM-Reference-Format}
\bibliography{sample-base}

\appendix

\end{document}